\documentclass{article}

\usepackage{PRIMEarxiv}

\usepackage[bookmarks]{hyperref}
\hypersetup{
	colorlinks,
	citecolor=black,
	filecolor=black,
	linkcolor=black,
	urlcolor=black,
	pdfauthor={},
	pdfsubject={},
	pdftitle={}
}
\usepackage{graphics} % for pdf, bitmapped graphics files
\usepackage{times} % assumes new font selection scheme installed
\usepackage{amsmath} % assumes amsmath package installed
\usepackage{amssymb}  % assumes amsmath package installed
\usepackage{graphicx}
\usepackage{algorithm}
\usepackage{lipsum}
\usepackage{orcidlink}
\usepackage{subcaption}
\usepackage{easyReview}
\usepackage{subcaption}
\usepackage{balance}
\usepackage[noend]{algpseudocode}
\usepackage{cite}
\usepackage{booktabs}
\usepackage{xcolor}
\def\etal{\emph{et al.}}

\def\secref#1{Sec.~\ref{#1}}
\def\figref#1{Fig.~\ref{#1}}
\def\tabref#1{Tab.~\ref{#1}}
\def\eqref#1{(\ref{#1})}

\newcommand\copyrighttext{%
	\small \begin{center} \color{red} \textcopyright \ 2025 IEEE. Personal use of this material is permitted. Permission from IEEE must be obtained for all other uses, in any current or future media, including reprinting/republishing this material for advertising or promotional purposes, creating new collective works, for resale or redistribution to servers or lists, or reuse of any copyrighted component of this work in other works. \end{center}}
\newcommand\copyrightnotice{%
	\begin{tikzpicture}[remember picture,overlay]
		\node[anchor=south,yshift=25.6cm] at (current page.south) 
		{\color{red}\fbox{\parbox{\dimexpr\textwidth-\fboxsep-\fboxrule\relax}{\copyrighttext}}};
	\end{tikzpicture}%
}

\title{\copyrightnotice \LARGE \bf Invisible Servoing: a Visual Servoing Approach with Return-Conditioned Latent Diffusion}

\author{{Bishoy Gerges \orcidlink{0000-0001-5066-3011}\thanks{Robotics and Mechatronics group, Faculty of Electrical Engineering, Mathematics and Computer Science, University of Twente, 7500AE Enschede, The Netherlands. \tt{ \footnotesize 
	\href{mailto:bishoy.y.gerges@gmail.com}{bishoy.y.gerges@gmail.com},
    \href{mailto:b.bazzana@utwente.nl}{b.bazzana@utwente.nl},
    \href{mailto:y.a.l.a.aboudorra@utwente.nl}{y.a.l.a.aboudorra@utwente.nl},
    \href{mailto:a.franchi@utwente.nl}{a.franchi@utwente.nl}
    }}} \And
Barbara Bazzana\orcidlink{0000-0002-2843-4324}$^{*}$ \And {Nicol\`o Botteghi \orcidlink{0000-0003-3676-3619}\thanks{Department of Applied Mathematics, University of Twente, 7500AE Enschede, The Netherlands. \tt{ \footnotesize 
	\href{mailto:n.botteghi@utwente.nl}{n.botteghi@utwente.nl}}}} \And Youssef Aboudorra\orcidlink{0000-0003-1896-2403}$^{*}$ \And {Antonio Franchi \orcidlink{0000-0002-5670-1282}$^{*,}$\thanks{Department of Computer, Control and Management Engineering, Sapienza University of Rome, 00185 Rome, Italy.
		{\tt \footnotesize
			\href{mailto:antonio.franchi@uniroma1.it}{antonio.franchi@uniroma1.it}
	} \\ This work was partially funded by the EU: AUTOASSESS project, EU Grant agreement ID: 101120732}}}

\begin{document}

\maketitle
%\thispagestyle{empty}
%\pagestyle{empty}

%%%%%%%%%%%%%%%%%%%%%%%%%%%%%%%%%%%%%%%%%%%%%%%%%%%%%%%%%%%%%%%%%%%%%%%%%%%%%%%%

\begin{abstract}
In this paper, we present a novel visual servoing (VS) approach based on latent Denoising Diffusion Probabilistic Models (DDPMs), that explores the application of generative models for vision-based navigation of UAVs (Uncrewed Aerial Vehicles). Opposite to classical VS methods, the proposed approach allows reaching the desired target view, even when the target is initially not visible. This is possible thanks to the learning of a latent representation that the DDPM uses for planning and a dataset of trajectories encompassing target-invisible initial views. A compact representation is learned from raw images using a Cross-Modal Variational Autoencoder. Given the current image, the DDPM generates trajectories in the latent space driving the robotic platform to the desired visual target. The approach has been validated in simulation using two generic multi-rotor UAVs (a quadrotor and a hexarotor). The results show that we can successfully reach the visual target, even if not visible in the initial view.  A video summary with simulations can be found in:
\href{https://youtu.be/2Hb3nkkcszE}{https://youtu.be/2Hb3nkkcszE}.
\end{abstract}

%%%%%%%%%%%%%%%%%%%%%%%%%%%%%%%%%%%%%%%%%%%%%%%%%%%%%%%%%%%%%%%%%%%%%%%%%%%%%%%%
\section{Introduction}
\label{sec:intro}

\begin{figure}[htbp]
        \centering
    \begin{subfigure}[c]{0.49\textwidth}
    \centering
        \includegraphics[width=1\textwidth]{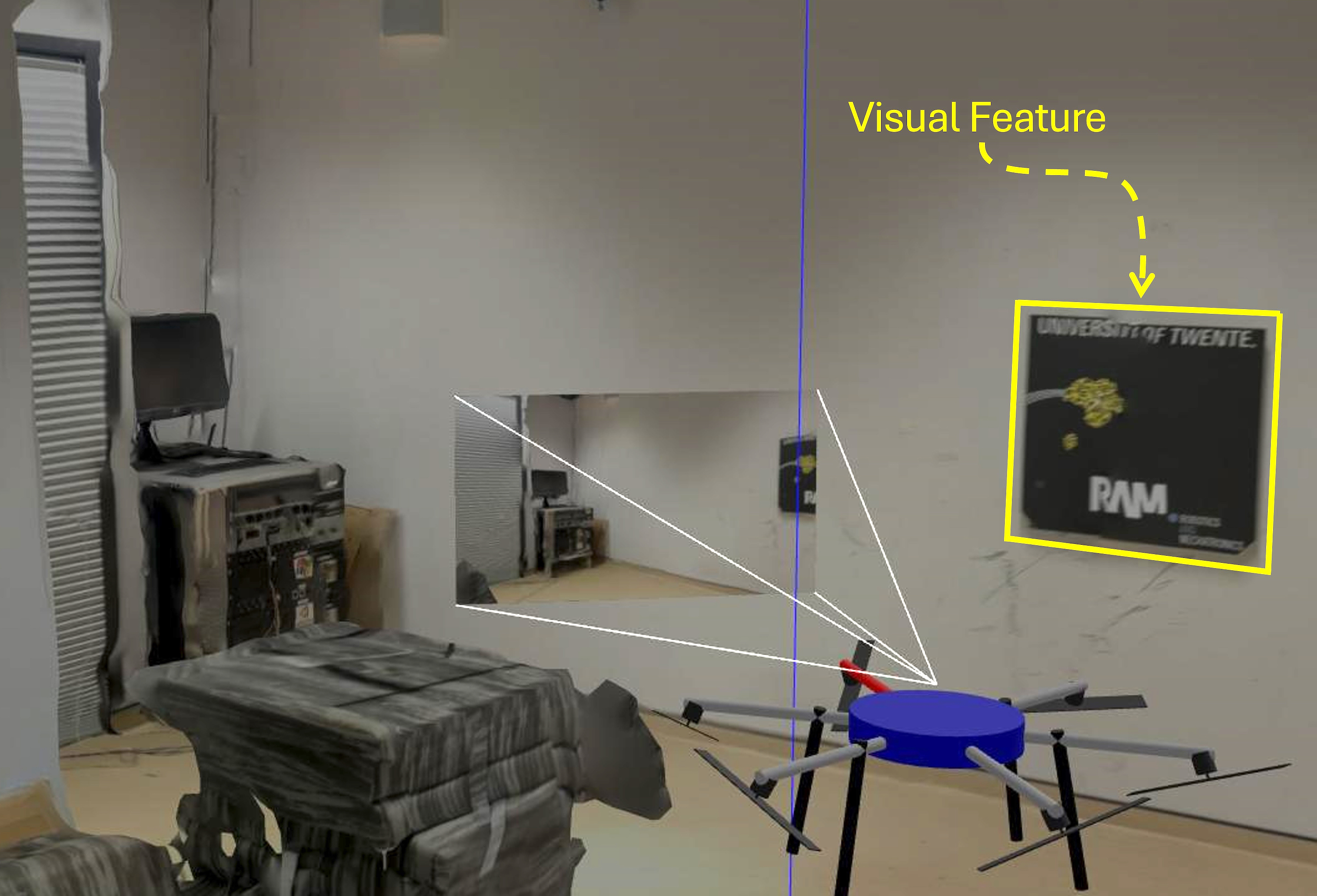}
    \caption{Gazebo simulation environment, camera view, and highlight of the visual target.}
    \end{subfigure}
    \label{pic:simtank}
        \begin{subfigure}[c]{0.49\textwidth}
           \centering
            \includegraphics[width=1\textwidth]{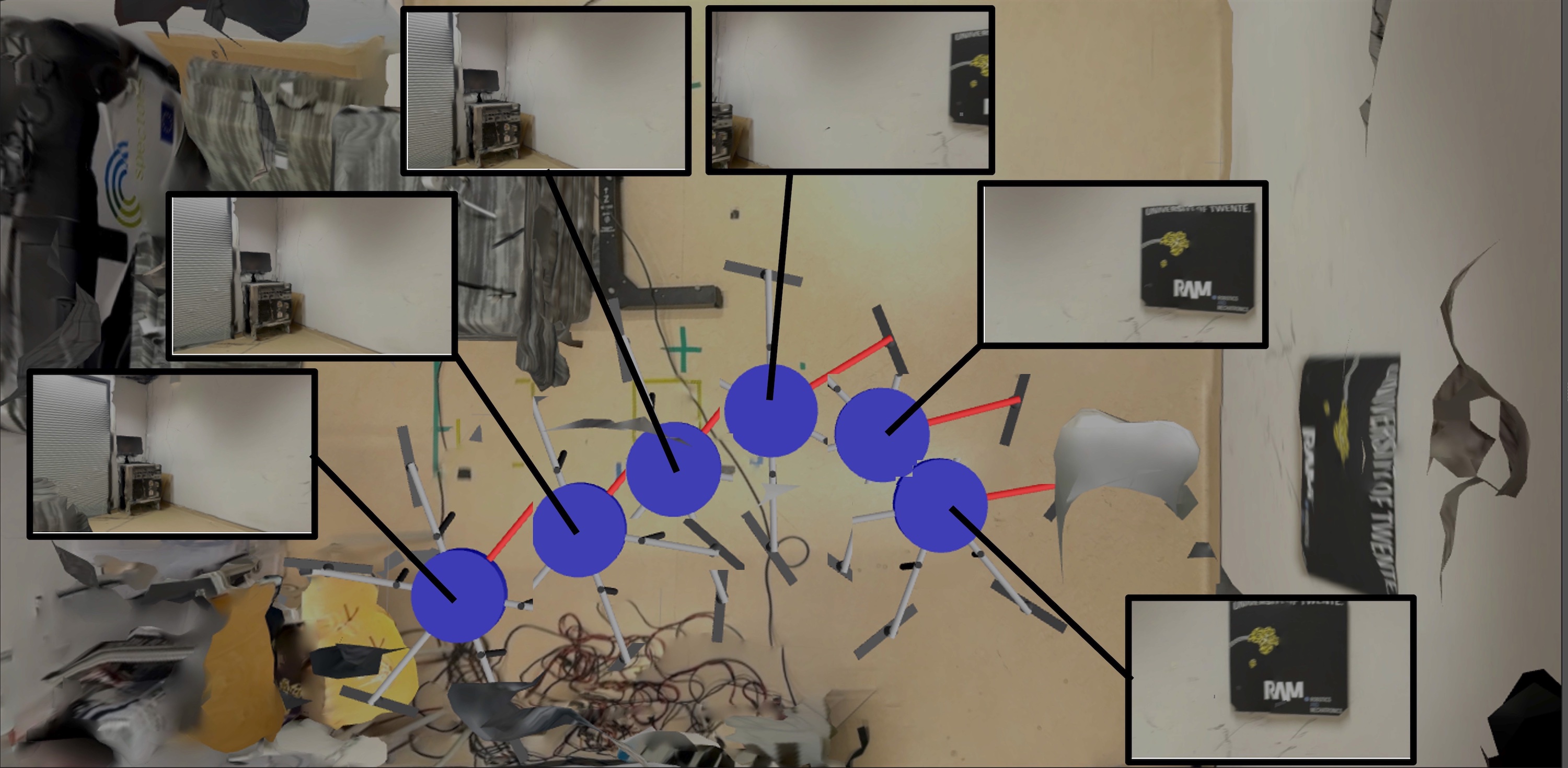}
    \caption{Five receding horizon steps of the returned-conditioned latent diffusion allow the UAV to reach the visual target.}
    \label{pic:simlab}
    \end{subfigure}
    \caption{An overview of the maneuver generated by the proposed approach to steer the UAV to the desired visual target, using the images of the onboard camera.}
    \label{fig:motivation}
    \vspace{-20px}
\end{figure}

Autonomous robotic solutions have been increasingly employed in industrial scenarios to protect human workers from dangerous and confined areas~\cite{uav_inspect1, ivic2023ac, bircher2015icra}. As an example, UAVs are used for inspecting safety-critical infrastructures, such as ballast tanks and cargo hulls. In industrial applications, an inspection task is commonly defined in terms of a desired target view in an environment known in advance. Such a task can be addressed using classical VS approaches~\cite{dvs1, dvs2}. However, these methods need to constantly keep visibility on the target view in order to define the error between the current and desired positions of the features.
This paper proposes a novel approach to VS using latent Denoising Diffusion Probabilistic Models (DDPMs)~\cite{diffusionHo, latentdiff}. Our approach leverages the capabilities of DDPMs in planning for control to perform VS that can handle the case of partial or complete visual target loss, as shown in~\figref{fig:motivation}. 

% \emph{HOW: Third, explain briefly HOW you do it.}\\
In particular, we employ DDPMs to perform VS planning in a lower-dimensional latent space. The lower-dimensional representation is obtained using a Cross-Modal Variational Autoencoder (CM-VAE)~\cite{cmvae, cmvae2}, that learns a regularized latent space, i.e., informative and smooth with respect to the VS task, shared between two data modalities: the image data and the relative position of the feature with respect to the robot. The VS task is then formulated as planning a trajectory of latent image frames that transitions from the initial view to the desired target view. This trajectory can be generated using DDPMs via \textit{inpainting}, i.e. producing the intermediate frames to fill the gap between initial and target frames. High-level control actions are then extracted from the generated trajectory via an inverse dynamics model~\cite{iscond}, which is a neural network that was trained alongside the DDPM to predict control actions given pairs of two consecutive states. The control actions are then fed to a motion controller that sends the control commands to the actuators of the robot. The control scheme works in a receding horizon manner, where only one or a limited number of the control actions are executed before a new planning iteration takes place. This approach is summarized in~\figref{fig:overall_arch}.

%\emph{Explain your contribution in one paragraph}\\
Our contribution goes as follows:
\begin{itemize}
    \item use of CM-VAE to learn a low-dimensional regularized latent space that is informative and smooth for the VS task;
    \item a return estimation algorithm that leverages the cross-modal regularized latent representation to estimate effective return values for conditional sampling with the DDPM;
    \item validation of the invisibility servoing in simulation using generic multi-rotor UAVs (\figref{fig:motivation}).
\end{itemize}
The approach can generalize to different robotic platforms, as the velocity actions planned by the DDPM are tackled by the specific motion controller. We have tested our approach in simulation, applying the planned actions in a receding horizon fashion to allow for online correction of model uncertainties. Future work includes experiments in real-world and more complex scenarios, and investigation of an end-to-end pipeline.

%%%%%%%%%%%%%%%%%%%%%%%%%%%%%%%%%%%%%%%%%%%%%%%%%%%%%%%%%%%%%%%%%%%%%%%%%%%%%%%%
\section{Related Work}
\label{sec:related}

% Classical VS that requre feature extraction
Classically, VS encompasses two primary methods: Image-based VS (IBVS) \cite{chaumette2006} and Position-based VS (PBVS) \cite{pbvs}. IBVS focuses on minimizing the error between the current and desired positions of features in the 2D image space. PBVS, on the other hand, operates by estimating the position of features relative to the camera and minimizing the error between the current and desired positions of features in the 3D space. The two methods can be combined to form a hybrid VS approach \cite{hvs}, where the error includes features from both the image and the 3D spaces. These methods require feature extraction leading to the main drawback of classical VS: the feature-loss problem. This occurs when the visual features used for guiding the robot become occluded, move out of the camera's field of view, or are otherwise undetectable due to changes in lighting, reflections, or other environmental conditions leading to a failure in the control loop. Jacquet~\etal~\cite{jacquetMpc} used perception-constrained nonlinear model predictive control (NMPC) to maintain visibility while tracking a trajectory with motor constraints. Similarly, Quin~\etal~\cite{mpcso3} applied MPC with spherical image-based servoing and visibility constraints. However, these approaches fail when the target is occluded in the initial image or lost during the execution of the trajectory due to external disturbances or lighting changes. Our approach instead can recover visibility on a user-defined target based on the experience learned from the dataset.

% DVS
An alternative to classical VS is Direct VS (DVS). DVS approaches operate directly on the visual input \cite{dvs1, dvs2}, eliminating the need for feature extraction and tracking. The visual input includes information like pixel intensity values or image gradients, which are used to determine the robot's control commands. By operating on the whole image, DVS formulates control laws based on the similarity between two images. Albeit novel approaches have been proposed to overcome the difficulty of a large-scale highly nonlinear problem, e.g. leveraging Discrete Orthogonal Moments~\cite{chen2024tro}, those approaches still require the visual target to be visible.

% DDPM
Recently, a new approach towards offline Deep Reinforcement Learning (DRL) in robotics has emerged exploiting DDPMs for planning and control \cite{janner2022, iscond}. By conditioning the generation process on suitable expected returns, the DDPM can strategically generate task-specific trajectories that maximize cumulative rewards over a prediction horizon. This has been demonstrated to even achieve better results than traditional offline DRL methods specially in tasks that require a high degree of flexibility and dexterity \cite{chi2023}. Leveraging the ideas of planning in the image space~\cite{martinez2021sj} and learning a latent representation of the images~\cite{felton2023icra}, this paper explores the use of DDPMs in the latent space for DVS. This has been possible building on the recent Cross-Modal Variational Autoencoders (VAEs) \cite{cmvae, cmvae2} and the return estimation algorithm we implemented.

\section{Background}

\subsection{Denoising Diffusion Probabilistic Models}

Denoising diffusion probabilistic models (DDPMs) \cite{Sohl-Dickstein2015DeepThermodynamics, Ho2020DenoisingModels} are a class of probabilistic generative models, such as generative adversarial networks \cite{Goodfellow2020GenerativeNetworks}, variational autoencoders \cite{Kingma2014Auto-encodingBayes}, and score-matching models \cite{Song2019GenerativeDistribution, Song2020ImprovedModels}. DDPMs are parametrized Markov chains that learn data distributions out of which new data can be generated from random noise by performing a step-wise denoising of the random vectors. The denoising process, often referred to as the reverse process, is learned from the dataset with the goal of reverting the forward process, which gradually adds noise to the original data $\mathbf{x}_t$ until convergence to a Gaussian distribution $\mathcal{N}(\mathbf{x}_t; \boldsymbol{\mu}, \mathbf{\Sigma})$ with zero mean $\boldsymbol{\mu}$ and unit variance $\mathbf{\Sigma}$. The forward process is detailed as follows:
\begin{equation}
\begin{split}
    q(\mathbf{x}_{1:T}|\mathbf{x}_0) &= \prod_{t=1}^Tq(\mathbf{x}_t|\mathbf{x}_{t-1})\, , \\   q(\mathbf{x}_t|\mathbf{x}_{t-1})&=\mathcal{N}(\mathbf{x}_t; \sqrt{1-\beta_t}\mathbf{x}_{t-1}, \beta_t\mathbf{I}) \, ,
\end{split}
\label{eq:forward_process}
\end{equation}
where we indicate with $\mathbf{x} \in \mathbb{R}^{|\mathbf{x}|}$  a generic data sample from a dataset $X$, the subscript $t=1, \dots, T$ indicates the $t^{\text{th}}$ diffusion step, $T$ the number of diffusion steps, $\mathbf{x}_0$ the noise-free data, and $\beta_t \in (0, 1)$ is a hyperparameter controlling the variance of the forward diffusion process. The reverse process can be written as in \eqref{eq:reverse_process}:
\begin{equation}
\begin{split}
    p_{\boldsymbol{\theta}}(\mathbf{x}_{0:T}) &= \prod_{t=1}^Tp_{\boldsymbol{\theta}}(\mathbf{x}_{t-1}|\mathbf{x}_t), \\ 
    p_{\boldsymbol{\theta}}(\mathbf{x}_{t-1}|\mathbf{x}_t&)=\mathcal{N}(\mathbf{x}_{t-1}; \boldsymbol{\mu}_{\boldsymbol\theta}(\mathbf{x}_t, t), \boldsymbol{\Sigma}_t)\, ,
\end{split}
    \label{eq:reverse_process}
\end{equation}
where $\boldsymbol{\mu}_{\boldsymbol\theta}(\mathbf{x}_t, t)$ is the mean learned by the DDPM, $\boldsymbol{\Sigma}_t$ is the covariance matrix, usually following a cosine schedule \cite{Nichol2021ImprovedModels}, i.e., not learned from data, and the subscript $\boldsymbol{\theta}$ indicates the learnable parameters of the diffusion model.

DDPMs can be trained by maximizing the variational lower bound similarly to variational autoencoders \cite{Sohl-Dickstein2015DeepThermodynamics}, or using the simplified objective introduced by \cite{Ho2020DenoisingModels}:
\begin{equation}
\begin{split}
\mathcal{L}(\boldsymbol{\theta}) &= \mathbb{E}_{t, \boldsymbol{\epsilon}, \mathbf{x}_0} = [||\boldsymbol{\epsilon} - \boldsymbol{\epsilon}_{\boldsymbol{\theta}}(\mathbf{x}_t, t) ||^2]\, , \\
&= \mathbb{E}_{t, \boldsymbol{\epsilon}, \mathbf{x}_0} = [||\boldsymbol{\epsilon} - \boldsymbol{\epsilon}_{\boldsymbol{\theta}}(\sqrt{\Bar{\alpha_t}}\mathbf{x}_0+\sqrt{1-\Bar{\alpha}_t}\boldsymbol{\epsilon}, t) ||^2]\, ,
\end{split}
    \label{eq:diffusion_simple_objective}
\end{equation}
where $t \sim \mathcal{U}\{1, \dots, T \}$ with $\mathcal{U}$ indicating a uniform distribution, $\boldsymbol{\epsilon} \sim \mathcal{N}(\mathbf{0}, I)$ is the target noise, $\Bar{\alpha}_t=\prod_{t=1}^T\alpha_t$ with $\alpha_t=1-\beta_t$, and with the learnable mean rewritten as:
\begin{equation}
\boldsymbol{\mu}_{\boldsymbol{\theta}}(\mathbf{x}_t, t)=\frac{1}{\sqrt{\alpha}_t}\Big(\mathbf{x}_t-\frac{1-\alpha_t}{\sqrt{1-\Bar{\alpha}_t}}\boldsymbol{\epsilon}_{\boldsymbol{\theta}}(\mathbf{x}_t, t)\Big).
    \label{eq:mu_rewritten}
\end{equation}

Conditional generation is the mechanism used by generative models to generate high-quality samples, where high-quality samples are, for example, samples belonging to a specified class label. Conditional generation for DDPMs can be done in two ways, namely the classifier-guidance \cite{Dhariwal2021DiffusionSynthesis} and the classifier-free \cite{ho2022classifier} generation. In the first case, we rely on an additional classifier to guide the sampling of the reverse process \eqref{eq:reverse_process}. However, this comes at the price of training an additional DDPM classifier that is capable of predicting the class labels from noise. Conversely, the classifier-free generation only relies on adding an additional input to the DDPM, i.e., the conditioning variable $\mathbf{c}$, to predict the noise level $\hat{\boldsymbol{\epsilon}}_t$ at a generic diffusion step $t$:
\begin{equation}
    \hat{\boldsymbol{\epsilon}}_t = \boldsymbol{\epsilon}_{\boldsymbol\theta}(\mathbf{x}_t, \varnothing, t) + \omega(\boldsymbol{\epsilon}_{\boldsymbol\theta}(\mathbf{x}_t, \mathbf{c}, t) - \boldsymbol{\epsilon}_{\boldsymbol\theta}(\mathbf{x}_t, \varnothing, t))\, ,
    \label{background:classifier_free:combined_score}
\end{equation}
where $\omega$ is a scalar coefficient chosen to balance the contribution of the unconditional generation via $ \boldsymbol{\epsilon}_{\boldsymbol\theta}(\mathbf{x}_t, \varnothing, t)=\boldsymbol{\epsilon}_{\boldsymbol\theta}(\mathbf{x}_t, \mathbf{c}=\varnothing, t)$, i.e., without conditioning variables, and the conditional generation via $\boldsymbol{\epsilon}_{\boldsymbol\theta}(\mathbf{x}_t, \mathbf{c}, t)$. Classifier-free DDPMs can be trained by modifying the objective in  \eqref{eq:diffusion_simple_objective}: 
\begin{equation}
\mathcal{L}(\boldsymbol{\theta}) = \mathbb{E}_{t, \boldsymbol{\epsilon}, \mathbf{x}_0, \delta \sim \text{Bern}(\pi)} [||\boldsymbol{\epsilon} - \boldsymbol{\epsilon}_{\boldsymbol{\theta}}(\mathbf{x}_t, (1-\delta)\mathbf{c} + \delta \varnothing, t) ||^2], \, 
    \label{eq:classifier-free_diffusion_simple_objective}
\end{equation}
where $\boldsymbol{\epsilon}_{\boldsymbol{\theta}}(\mathbf{x}_t, (1-\delta)\mathbf{c} + \delta \varnothing, t) = \boldsymbol{\epsilon}_{\boldsymbol{\theta}}(\sqrt{\Bar{\alpha_t}}\mathbf{x}_0+\sqrt{1-\Bar{\alpha}_t}\boldsymbol{\epsilon}, (1-\delta)\mathbf{c} + \delta \varnothing, t)$, $\delta \in \{0, 1\}$ is sampled from a Bernoulli distribution $\text{Bern}(\pi)$ of parameter $\pi$ and determines whether we predict $\boldsymbol{\epsilon}$ in unconditioned settings, i.e., for $\delta=1$ we recover  \eqref{eq:diffusion_simple_objective}, or in conditioned settings, i.e. for $\delta=0$.

\begin{figure}[t]
  \centering
 \includegraphics[width=0.7\columnwidth]{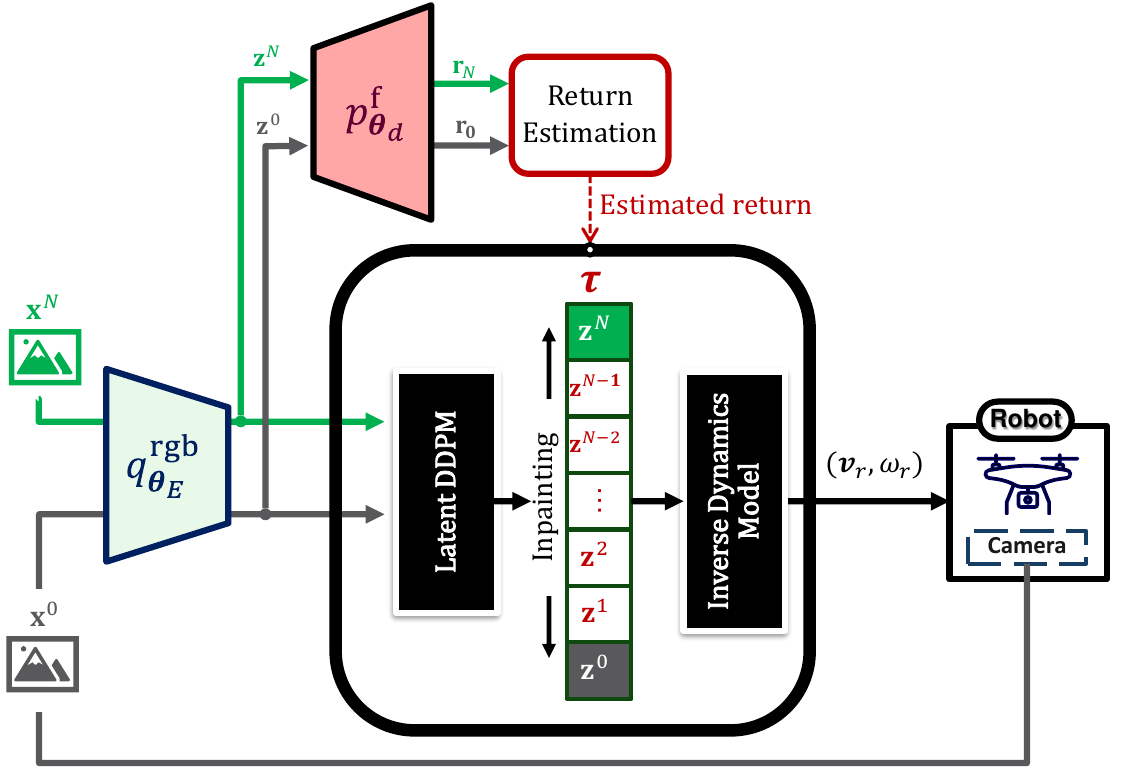}
  \caption{Architecture of the latent diffusion VS approach.}
  \label{fig:overall_arch}
  \vspace{-10px}
\end{figure}

\subsection{Planning and Control with DDPMs}

By reconceptualizing the input data, the use of DDPMs can be extended from image generation to strategic planning for control. In this approach, a 2D image array is interpreted not as pixels, but as a 2D array representing system states and actions along a control horizon $N$. The columns of the array represent state-action pairs of the system at different time instants along the decision-making horizon ~\cite{janner2022, iscond}. 
Consider a dataset $\mathcal{D}$ composed of $M$ trajectories $\boldsymbol{\tau}$ defined as:
\begin{equation}
    \boldsymbol{\tau} = \begin{bmatrix}
    \mathbf{s}^0 & \ldots &\mathbf{s}^k & \ldots & \mathbf{s}^N \\
    \mathbf{a}^0 & \ldots &\mathbf{a}^k & \ldots & \mathbf{a}^N
    \end{bmatrix} \,,
\end{equation}
where $\mathbf{s}$ is the state vector of the system, $\mathbf{a}$ the control input, $N$ is the decision-making (prediction) horizon, and $k$ is a generic instant of time in that horizon. 
%DDPMs can leverage a dataset of trajectories to learn the system dynamics. 
The DDPM can then be trained to approximate the distribution of the system dynamics corresponding to the distribution of the trajectories in the dataset. Once trained, the DDPM can then generate new trajectories that adhere to the learned dynamics distribution, employing the same principles used for handling image data.

As shown in \cite{janner2022}, the generation task can be simplified by generating state trajectories and using an inverse dynamics model to recover the control actions that steer the system from one state to the next \cite{iscond}. The inverse dynamics model, $f_{\boldsymbol{\phi}}$ of parameters $\boldsymbol{\phi}$, is trained by minimizing the expected mean squared error between the predicted control action and the actual control action in the training dataset. The loss function is defined as:
\begin{equation}
    \mathcal{L}_({\boldsymbol{\phi}}) = \mathbb{E}_{\mathbf{s}^k, \mathbf{s}^{k+1}, \mathbf{a}^k \sim \mathcal{D}}[||f_\phi(\mathbf{s}^k, \mathbf{s}^{k+1}) - \mathbf{a}^k||^2] \,,
\label{eq:inverse_model_loss}
\end{equation}
where $\mathbf{s}^k$ and $\mathbf{s}^{k+1}$ are the state vectors at time $k$ and $k+1$ respectively, $\mathbf{a}^k$ is the control action at time $k$. The inverse dynamics model is trained simultaneously with the diffusion model by adding its loss function to the loss function of the diffusion model in \eqref{eq:classifier-free_diffusion_simple_objective}.\\

To generate optimal trajectories with respect to a task-based reward (or cost) function, we can exploit conditional sampling. Conditional sampling allows for the generation of state trajectories that do not just comply with the learned dynamics but can be specifically conditioned on various factors such as desired behaviors, starting and/or end states, or, notably, expected returns\footnote{The expected return corresponds to the expected discounted cumulative reward over the planning horizon $N$.}.  Return-conditioned diffusion has been shown \cite{janner2022, iscond, chi2023} to outperform offline DRL methods in strategically generating task-specific trajectories.\\

To generate optimal trajectories with respect to a task-based reward (or cost) function, we can exploit conditional sampling. Conditional sampling allows for the generation of state trajectories that do not just comply with the learned dynamics but can be specifically conditioned on various factors such as desired behaviors, starting and/or end states, or, notably, expected returns\footnote{The expected return corresponds to the expected discounted cumulative reward over the planning horizon $N$.}.  Return-conditioned diffusion has been shown \cite{janner2022, iscond, chi2023} to outperform offline DRL methods in strategically generating task-specific trajectories.\\

%%%%%%%%%%%%%%%%%%%%%%%%%%%%%%%%%%%%%%%%%%%%%%%%%%%%%%%%%%%%%%%%%%%%%%%%%%%%%%%%
\section{Methodology}
\label{sec:main}

Building upon the planning for control concepts using DDPMs discussed in the previous section, a novel framework for VS is proposed (see \figref{fig:overall_arch}). In this approach, the system state is the image frame captured by the robot's camera. The VS task can be formulated as planning a trajectory of image frames that transitions from the initial frame $\mathbf{x}^0$ to the desired target frame $\mathbf{x}^N$. This trajectory can be generated using DDPMs by conditioning on the initial and target frames. In this context, the diffusion model fills the gap between the two by generating the intermediate frames. This process is called \textit{inpainting}\footnote{We borrow this terminology from the context of image generation where part of the image pixels is erased and the model is asked to fill in the missing pixels \cite{inpainting}.}. In this case, the target image is an image of a specific visual feature within the environment that the robot needs to reach. The dataset used to train the diffusion model is a set of different trajectories of image frames captured by the robot's camera within the environment. Each trajectory is generated by randomly spawning the robot at a random position and orientation in the environment and then tasking it to reach and observe a specified target. The set of
images contains frames with and without the visual feature,
which is captured from different angles.

\begin{figure}[t]
    \centering
    \includegraphics[width=0.7\columnwidth]{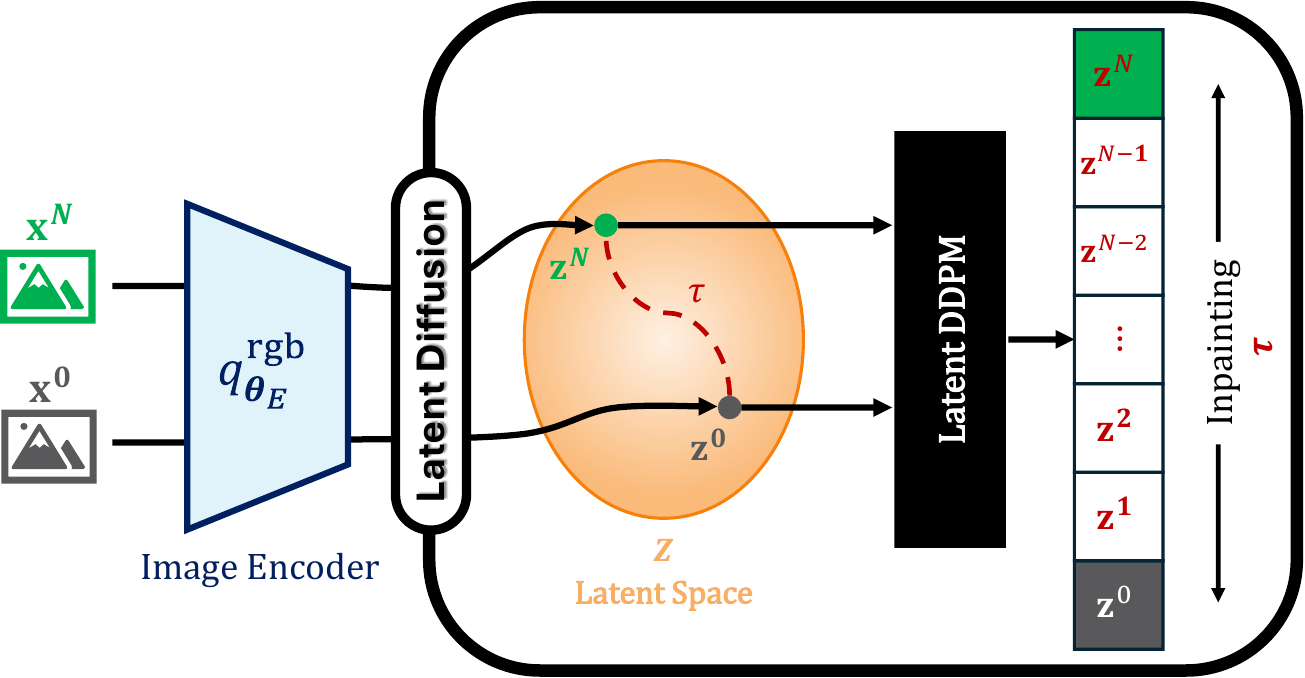}
    \caption{VS with latent DDPMs. We encode the initial frame $\mathbf{x}^0$ and the target frame $\mathbf{x}^N$ using the CM-VAE encoder. Their respective latent representations $\mathbf{z}^0$ and $\mathbf{z}^N$ are inpainted as initial and target latent states of the trajectory generated by the latent DDPM.}
    \label{fig:latent_ddpm}
\end{figure}

However, the high dimensionality of image data makes the methods introduced in \cite{janner2022, iscond} not directly applicable due to the need for generating extremely high-dimensional state trajectories, which is computationally untractable. Our novel framework builds upon the observation that the input data, i.e., the camera images, contains redundant information and that the relevant one lies in a low-dimensional space. Therefore, we can simplify the problem of VS with DDPM by taking inspiration from the state-of-the-art latent DDPM \cite{rombach2022high}. 
% Hence, visual servoing can be done in the latent space of the image data, where instead of using the image pixels as the system state, its latent vector representation $\mathbf{z}$
The problem of VS can then be solved in two steps: (i) learn a low-dimensional representation of the input data using a Cross-Modal Variational Autoencoder, and (ii) learn to generate latent space trajectories using a DDPM (see \figref{fig:latent_ddpm}).

\subsection{Dimensionality Reduction using Cross-Modal Variational Autoencoder}

To address the challenge of high dimensionality in image data and learn a compact and regularized latent space, a Cross-modal Variational Autoencoder (CM-VAE) \cite{cmvae} is utilized. This approach has been shown to reduce the data dimensionality effectively, while ensuring the generated latent space is smooth, continuous, consistent, and robust against variations in visual information \cite{cmvae, cmvae2}. The idea behind the CM-VAE is to integrate multiple data modalities into a single latent space. The CM-VAE processes two primary modalities: (i) RGB image data $\mathbf{x}_{\text{rgb}}$ and (ii) the visual feature's relative pose $\mathbf{x}_{\text{f}}$ with respect to the robot's coordinate frame, i.e., $\mathbf{x}_{\text{f}}=[r,\theta,\phi,\gamma]$, expressed using the spherical coordinates and the yaw. 

The input image $\mathbf{x}_{\text{rgb}}$ is projected by a deep neural network-based encoder $q^{\text{rgb}}_{\boldsymbol{\theta}_E}$ of parameters $\boldsymbol{\theta}$ into a normal distribution $\mathcal{N}(\boldsymbol{\theta};\boldsymbol{\mu}, \boldsymbol{\Sigma})$, from which the latent vector $\mathbf{z}$ is sampled\footnote{Similarly to classical VAEs, the encoder learns the mean and covariance of the Gaussian distribution over the latent variables. To sample from such a distribution a latent vector $\mathbf{z}$ the reparametrization trick is used \cite{Kingma2014Auto-encodingBayes}.}. The latent vector $\mathbf{z}$ is used to reconstruct each data modality by two dedicated deep neural network-based decoders $p^{\text{rgb}}_{\boldsymbol{\theta}_D}$ for images and $p^{\text{f}}_{\boldsymbol{\theta}_d}$ for feature poses.
More specifically, the full latent vector is used in the decoding of the image data, while for decoding the feature relative pose only the first four scalars of the latent vector are used, i.e., $\tilde{\mathbf{z}} = [z_1, z_2, z_3, z_4]$ is decoded to $r, \theta, \phi$ and $\gamma$ respectively. For the encoder and the two decoders we utilize the same neural network architecture introduced in \cite{cmvae}.

The CM-VAE is trained in an analogous way to a VAE \cite{Kingma2014Auto-encodingBayes} by maximizing the log likelihood (reconstruction terms) of image $\mathbf{x}_{\text{rgb}}$ and coordinate frame $\mathbf{x}_{\text{f}}$ given the latent vector $\mathbf{z}$ and $\tilde{\mathbf{z}}$, respectively, while minimizing the KL divergence (regularization term) between the encoder's output distribution $q_\text{rgb}(\textbf{z}|\textbf{x}_\text{rgb})$ and a standard normal distribution $\mathcal{N}(\mathbf{0}, I)$, representing the prior distribution of the latent vectors $p(\mathbf{z})$:
\begin{equation}
\begin{split}
      \mathcal{L}(\boldsymbol{\theta}_E, \boldsymbol{\theta}_D, \boldsymbol{\theta_d}) &= \underbrace{\mathbb{E}_{q^{\text{rgb}}_{\boldsymbol{\theta}_E}}[\log p^{\text{rgb}}_{\boldsymbol{\theta}_D}(\textbf{x}_\text{rgb}|\textbf{z})] + \mathbb{E}_{q_{\text{rgb}}}[\log p^{\text{f}}_{\boldsymbol{\theta}_d}(\textbf{x}_\text{f}|\tilde{\textbf{z}})]}_{\text{Reconstruction terms}} \\ &- \underbrace{D_{KL}(q_\text{rgb}(\textbf{z}|\textbf{x}_\text{rgb}) \| p(\textbf{z}))}_{\text{Regularization term}}.  
\end{split}
    \label{eq:cmvae_elbo}
\end{equation}

\subsection{Latent DDPMs for Visual Servoing}
As shown in \figref{fig:latent_ddpm}, the image encoder $q^{\text{rgb}}_{\boldsymbol{\theta}_E}$, trained to minimize the CM-VAE loss function in \eqref{eq:cmvae_elbo},  is used to encode the image data into the regularized latent space for the diffusion model to operate in. In particular, the DDPM task is now to learn to generate latent state trajectories of the type:
\begin{equation}
    \boldsymbol{\tau} = \begin{bmatrix}
    \mathbf{z}^0 & \ldots &\mathbf{z}^k & \ldots & \mathbf{z}^N \\
    \end{bmatrix} \,,
\end{equation}
where the control actions $\hat{\mathbf{a}}$ are recovered using an inverse model $f_{\boldsymbol{\phi}}$ taking as input consecutive latent state pairs, e.g., $\hat{\mathbf{a}}^k=f_{\boldsymbol{\phi}}(\mathbf{z}^k, \mathbf{z}^{k+1})$.
In particular, we choose high-level control actions, namely linear and angular velocities, because they are platform-agnostic.

% Training the diffusion model
To train the DDPM model, we project the image trajectories into the latent space using the trained CM-VAE to obtain the trajectories in the latent space. For conditional sampling, the diffusion model is conditioned on a desired outcome or quality expressed as a desired return to achieve. To make this possible, the dataset is augmented with the corresponding returns achieved by each trajectory. Then, the latent DDPM is trained by a straightforward modification of classifier-free guidance objective introduced in~\eqref{eq:classifier-free_diffusion_simple_objective}, jointly with the latent inverse model (see~\eqref{eq:inverse_model_loss}):
\begin{equation}
\begin{split}
    &\mathcal{L}(\boldsymbol{\theta}, \boldsymbol{\phi}) = \\ 
    &\mathbb{E}_{t, \boldsymbol{\epsilon}, \mathbf{z}_0, \delta \sim \text{Bern}(\pi)} [||\boldsymbol{\epsilon} - \boldsymbol{\epsilon}_{\boldsymbol{\theta}}(\mathbf{z}_t, (1-\delta)R + \delta \varnothing, t) ||^2] + \\
    &\mathbb{E}_{\mathbf{z}^k, \mathbf{z}^{k+1}, \mathbf{a}^k\sim\mathcal{D}}[||f_\phi(\mathbf{z}^k, \mathbf{z}^{k+1}) - \mathbf{a}^k||^2], \,
\end{split}
    \label{eq:latent_ddpm_objective}
\end{equation}
where the conditioning variable $R$ corresponds to the expected return of the generated trajectory:
\begin{equation}
    R = -\sum_{k=0}^{N} \gamma^k r^k \,,
\end{equation}
and $r^k$  and $\gamma$ indicate a task-dependent instantaneous reward at time step $k$ and the discount factor, respectively.

\begin{figure}[t]
        \centering
        \includegraphics[width=0.4\columnwidth]{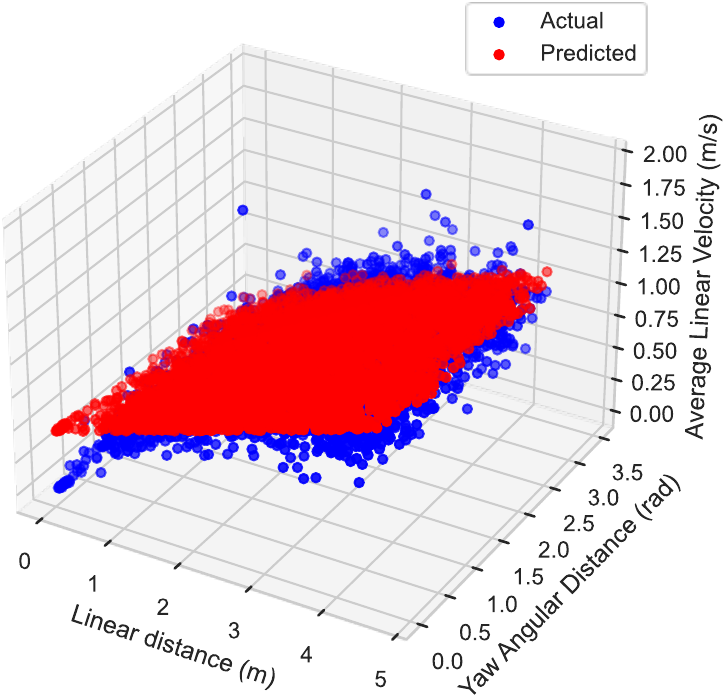}
        \caption{Linear regression model prediction of the velocity.}
        \label{fig:ret-est}
       %\vspace{-20px}
\end{figure}

\subsubsection{Return Estimation}
\label{sec:ret-estimation}
\begin{figure*}[t]
    \centering
    \includegraphics[width=0.85\textwidth]{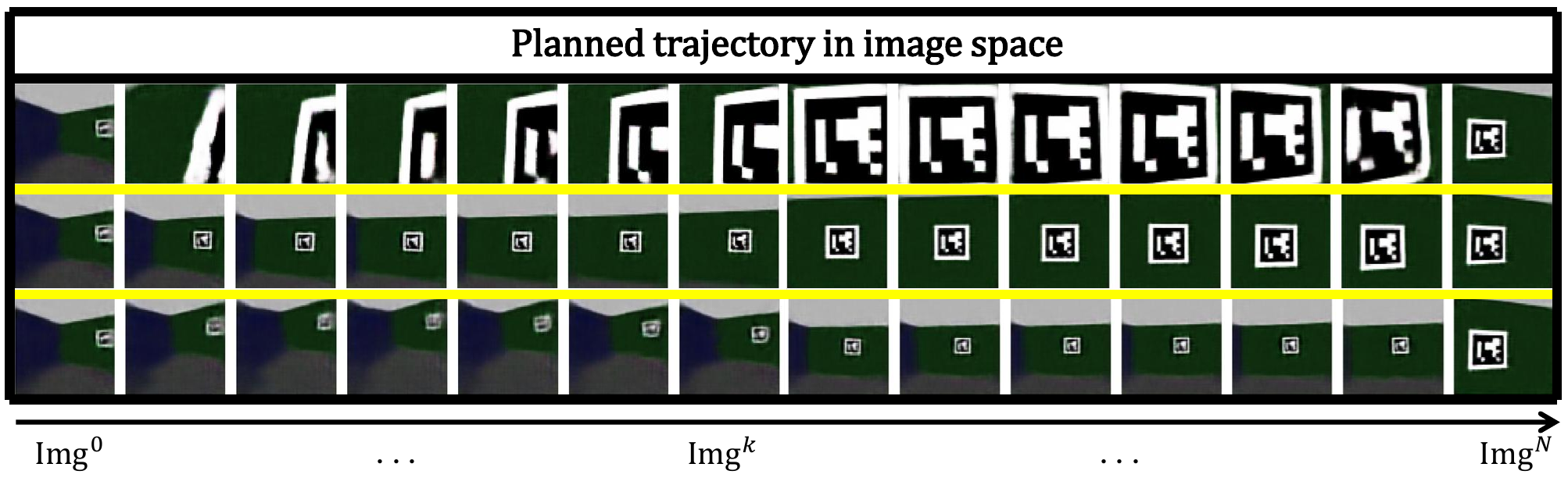}
    \caption{Effect on the quality of the planned trajectory of the return $R$ used for conditioning. Top: $R=1.0$; middle: $R=0.4$; bottom: $R=-1.0$.} 
    \label{fig:return_swap}
\end{figure*}

When conditionally generating trajectories, the choice of the return value $R$ is essential for the quality of the planning. In particular, a too large value for the return will result in physically-inconsistent trajectories with unachievable control requirements, while too small value will result in over-conservative trajectories, as shown in~\figref{fig:return_swap}. However, selecting the correct return value is not trivial.

In our VS task, the diffusion model is trained on a dataset in which the reward at any step in the trajectory is equal to minus the distance between the visual feature and the robot. Therefore, during planning, if the distance between the visual feature and the robot is known, the return can be estimated by assuming a constant-velocity motion of the robot from the start to the target image.
The position vector between the visual feature and the robot at the start and end of the trajectory can be calculated using the feature decoder $p^\text{f}_{\boldsymbol{\theta}_d}$ of the CM-VAE, which takes the encoded image latent vector $\textbf{z}$ as input and outputs the vector $[r,\theta,\phi,\gamma]$. The spherical coordinates $[r,\theta,\phi]$ can be converted to Cartesian coordinates to get the position vectors $\mathbf{r}_0$ and $\mathbf{r}_N$ at the start and end of the trajectory respectively, where $N$ is the length of the prediction horizon. The distance vector $\mathbf{d}$ between the start $\mathbf{r}_0$ and target position $\mathbf{r}_N$ can be calculated as $ \mathbf{d} = \mathbf{r}_0 - \mathbf{r}_N$.
Assuming that the robot moves with a constant velocity of magnitude $v$ along this vector $\mathbf{d}$ from the start position to the target position, the total travel time can be computed as $T=||\mathbf{d}||/v$.
Therefore, the time between each step $k$ in the trajectory can be calculated as $\Delta t = T / N$.
The position vector $\mathbf{r}_k$ between the visual feature and the robot at each step $k$ in the trajectory can be computed by sliding along the vector $\mathbf{d}$ with a step size of $v\Delta t$ starting from the start position $\mathbf{r}_0$:
\begin{align}
    \mathbf{r}_k   &= \left(\mathbf{d}-vk\Delta t\mathbf{\hat{d}}\right)+\mathbf{r}_N\\
                &= \left(||\mathbf{d}||-vk\Delta t\right)\mathbf{\hat{d}}+\mathbf{r}_N \,,
\end{align}
where $\mathbf{\hat{d}}$ is the unit vector in the direction of $\mathbf{d}$. The return is then estimated by summing the discounted rewards along the trajectory where the reward at each step $k$ is the negative of the distance $||\mathbf{r}_k||$ between the visual feature and the robot at that step:
\begin{equation}
    \text{R}_{est} = -\sum_{k=0}^{N} \gamma^k ||\mathbf{r}_k||.
\end{equation}

The constant velocity magnitude $v$ in the return estimation algorithm affects the calculated return. Therefore, the suitable value of $v$ should be chosen based on the robot's dynamics and the environment. 
To estimate this velocity, the dataset is used to calculate this velocity with two different approaches. The first approach is to calculate the average velocity of the robot in the dataset and use this velocity for any planning regardless of the start and target images. The second approach is to predict the velocity based on the linear and yaw angular distances between the start and target images where, again, these are calculated using the feature decoder of the CM-VAE. This velocity prediction is obtained from a linear regression model (linear 3D plane) that has been fit to the dataset as shown in~\figref{fig:ret-est}. The model takes the linear and yaw angular distances as input and outputs the predicted velocity.
The two approaches to select the velocity magnitude $v$ yield very similar results.

\begin{figure}[t]
        \centering
    \begin{subfigure}{0.45\columnwidth}
        \includegraphics[width=1\columnwidth]{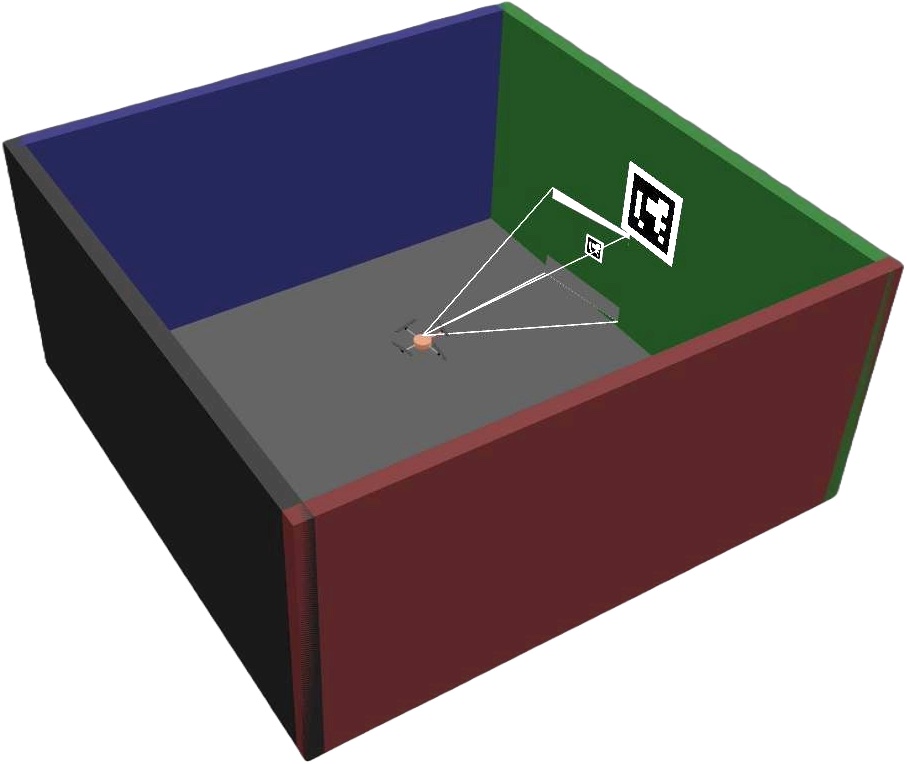}
    \caption{Box simulation environment of dimension $5\,\mathrm{m}^3$.}
    \label{pic:tank}
    \end{subfigure}
           \begin{subfigure}{0.50\columnwidth}
                \includegraphics[width=1\columnwidth]{img/lab_tx_1_ann.jpg}
    \caption{RAM Aerial Robotics Lab simulation environment.}
    \label{pic:lab}
    \end{subfigure}
    \caption{Testing environments.}
\end{figure}

\begin{figure}[t]
        \centering
        \includegraphics[width=0.7\columnwidth]{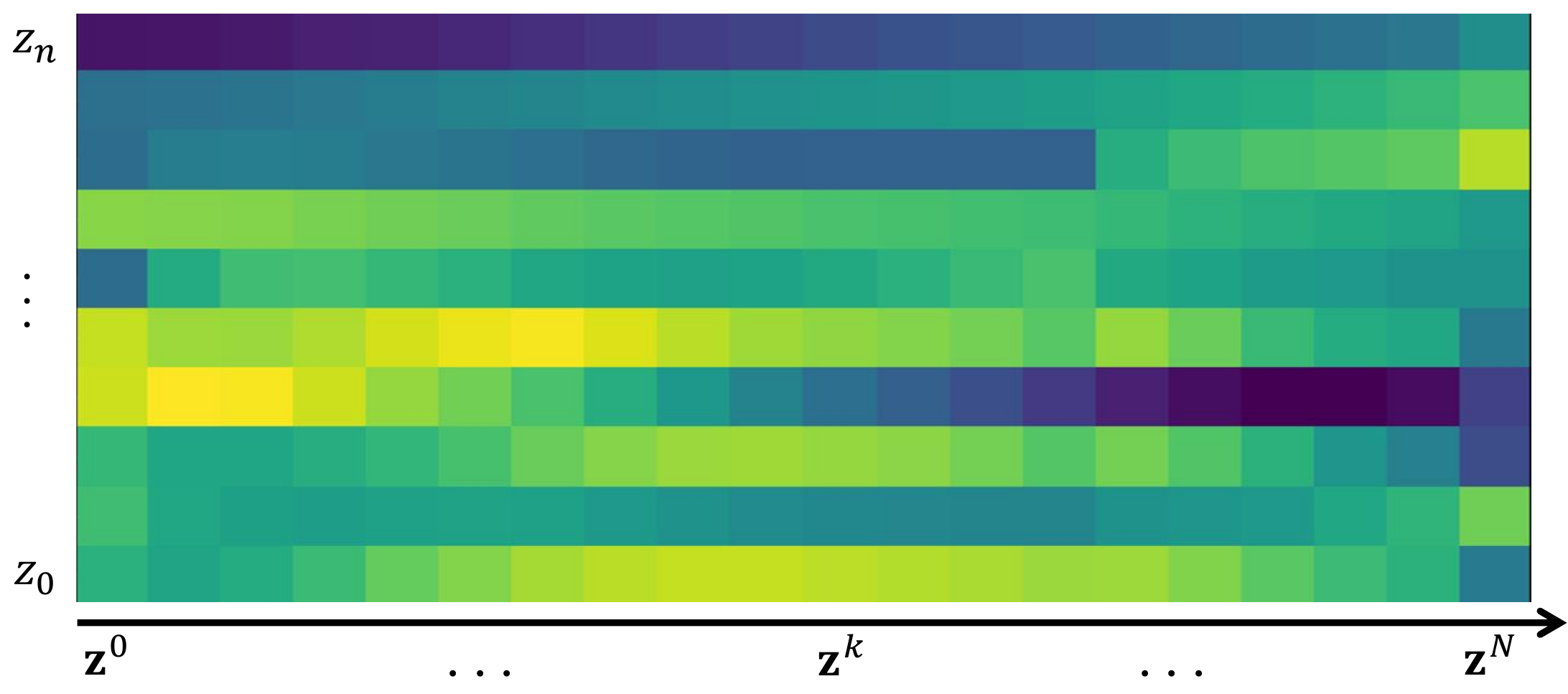}
        \caption{Planned trajectory in the latent space: each column corresponds to a time instant in the horizon.}
        \label{fig:ex-latent}
       %\vspace{-20px}
\end{figure}

%%%%%%%%%%%%%%%%%%%%%%%%%%%%%%%%%%%%%%%%%%%%%%%%%%%%%%%%%%%%%%%%%%%%%%%%%%%%%%%%
\section{Simulations}
\label{sec:exp}
The main focus of this work is to propose a novel VS approach based on latent DDPMs, that leverages a cross-modal regularized latent representation to estimate effective return values for conditional sampling. In particular, we want to test the capability of the DDPM-based controller to recover visibility and approach a given target view. Therefore, the simulation environments encompass a visual target, that the robot has to reach being spawned at random positions in the rooms.

To test our approach, we have considered two experimental scenarios: a simplified simulated environment of an empty square room (\figref{pic:tank}), and the Aerial Robotics Lab at our department (\figref{pic:lab}). In the former, the camera resolution is $128\times 128$ pixels and the FoV $1.21\,\mathrm{rad}$. The dataset consists of quadrotor trajectories, counting up to 700k image-datapoint pairs and containing: captured images, the visual feature pose w.r.t the robot, the reward at each time step, the robot's linear and angular velocities used as high-level control inputs. The visual target is the Aruco tag visible in~\figref{pic:tank}. 
In the simulated Aerial Robotics Lab, the camera resolution is $320\times 180$ pixels, and the FoV $1.21\,\mathrm{rad}$. The dataset consists of hexarotor and quadrotor trajectories, counting up to 700k image-datapoint pairs. The visual target is the RAM logo attached to the wall, visible in~\figref{pic:lab}.
In the datasets, the return values are normalized between -1.0 and 1.0.

The Gazebo simulations show the capabilities of our method to reach a desired target view, even if this is not visible at the start. The evaluation focuses on (i) the performance of the return estimation heuristic built upon the latent representation given by the CM-VAE, (ii) the performance of the diffusion-based controller working in a receding horizon fashion. 

\subsection{Return estimation}
As the dynamics are not modeled as hard constraints by the proposed approach, the effectiveness of the return estimation module is essential to generate dynamically feasible trajectories. Once a suitable return value has been calculated, planning takes place in the latent space. For illustration purposes, a color map of a generated latent trajectory is shown in~\figref{fig:ex-latent}. Once a latent space trajectory has been generated similar to the one in~\figref{fig:ex-latent}, the decoder $p^{\text{rgb}}_{\boldsymbol{\theta}_D}$ is used to visualize the trajectory in the image space converting each latent vector to an image.

\begin{figure}[t]
    \centering
    \includegraphics[width=0.7\columnwidth]{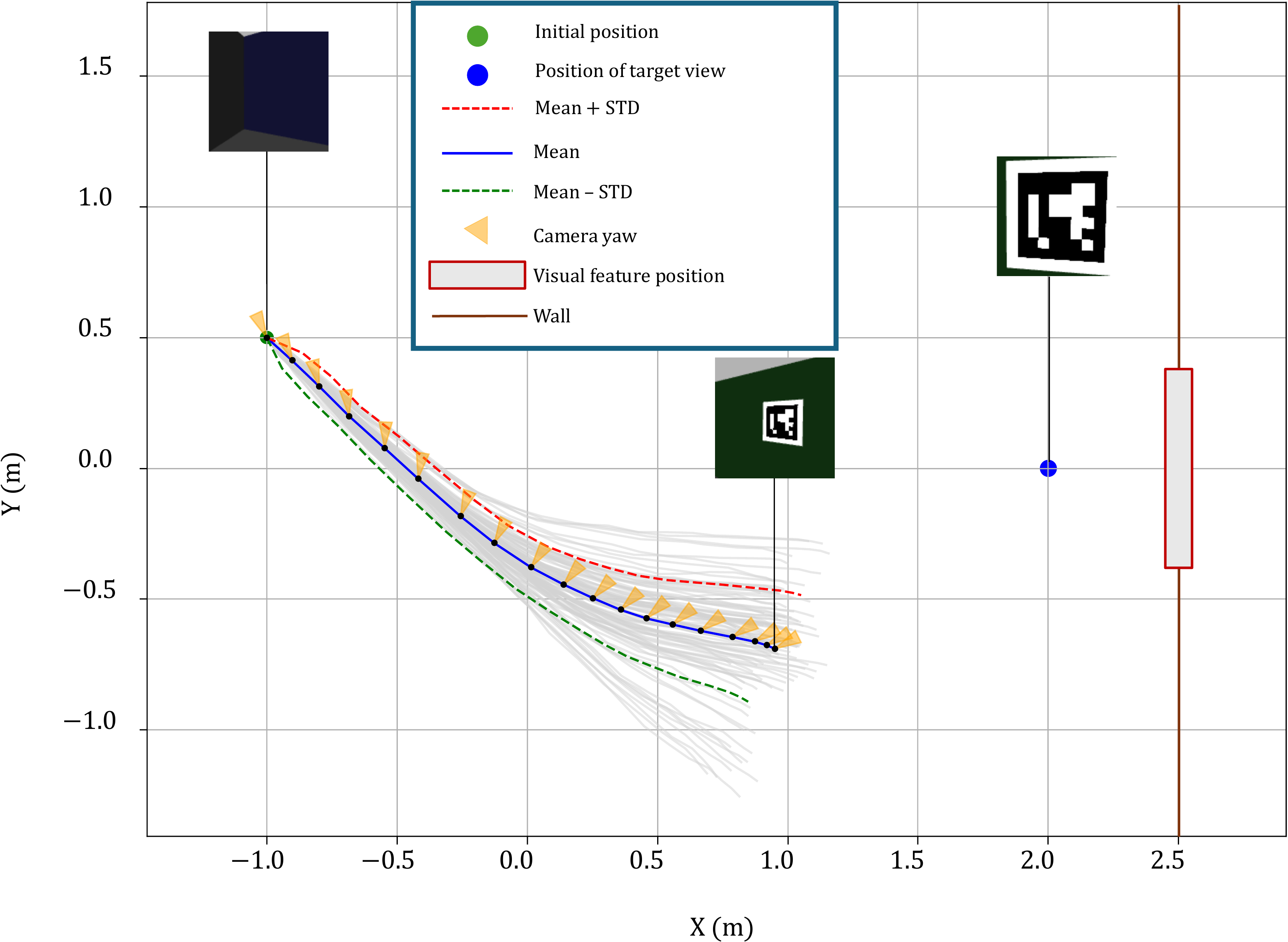}
    \caption{100 samples of generated UAV trajectories in open-loop fashion, along with their mean and standard deviation. The visual target is not visible from the initial location (-1.0, 0.5). The latent DDPM is capable of recovering the view of the target.} 
    \label{fig:open_loop}
\end{figure}

\figref{fig:return_swap} shows the effect of the return value used for conditioning. Because the architecture does not involve hard constraints, when the return value is too high, the trajectory jumps to the target to maximize the return, i.e., minimizing the discounted sum of the relative distance to the target. On the contrary, too small returns result in over-conservative trajectories that keep the UAV far from the target.

This analysis clarifies the importance of the return estimation module.
To validate it, we verified that the estimated return falls within the interval of admissible values, by doing a sweep analysis in $[-1.0,\,1.0]$ with resolution 0.2 for a given initial and target images. In a challenging scenario where the target is not visible, a sweep analysis has shown that the interval of admissible values resulting in a planned trajectory that smoothly converges to the target image without abrupt jumps is $[0.2,\,0.4]$. The return estimated by the algorithm described in~\secref{sec:ret-estimation} is 0.34, which is consistent with the result of the sweep analysis.

\begin{figure*}
    \centering

    \begin{minipage}{0.48\textwidth}
        \centering
        \includegraphics[width=\textwidth]{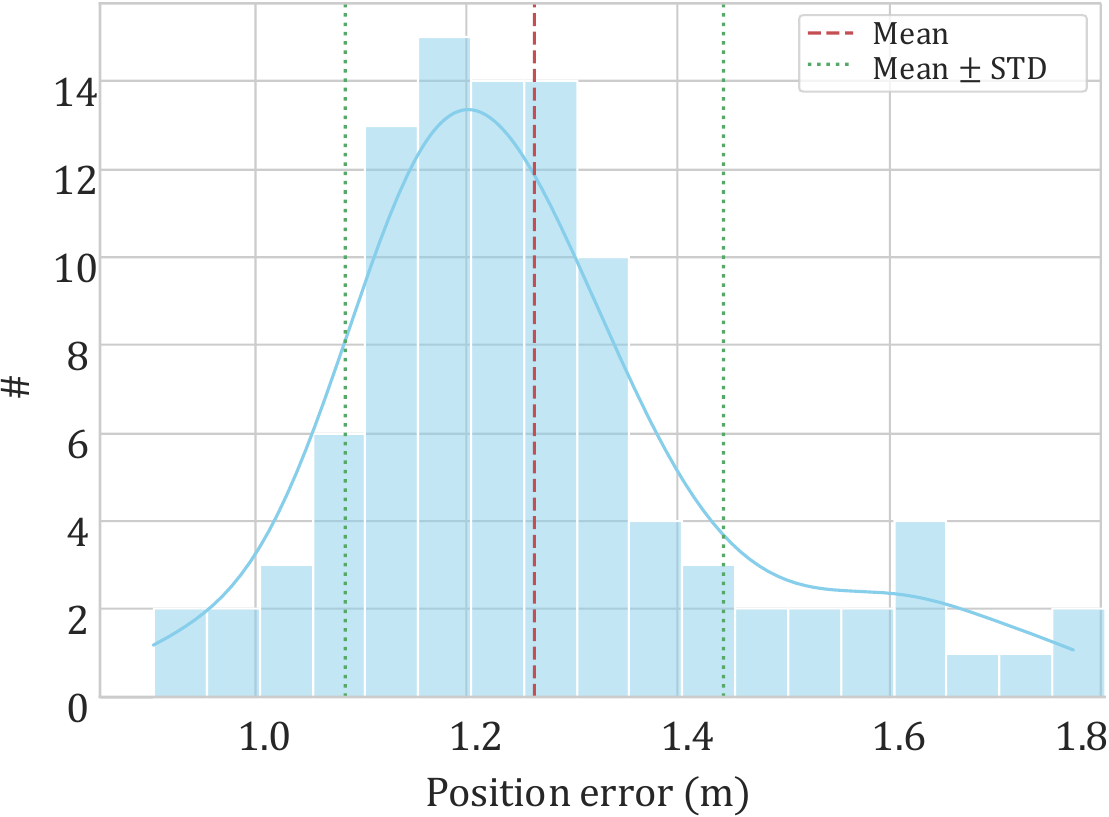}
        \\ (A)
    \end{minipage}
    \begin{minipage}{0.48\textwidth}
        \centering
        \includegraphics[width=\textwidth]{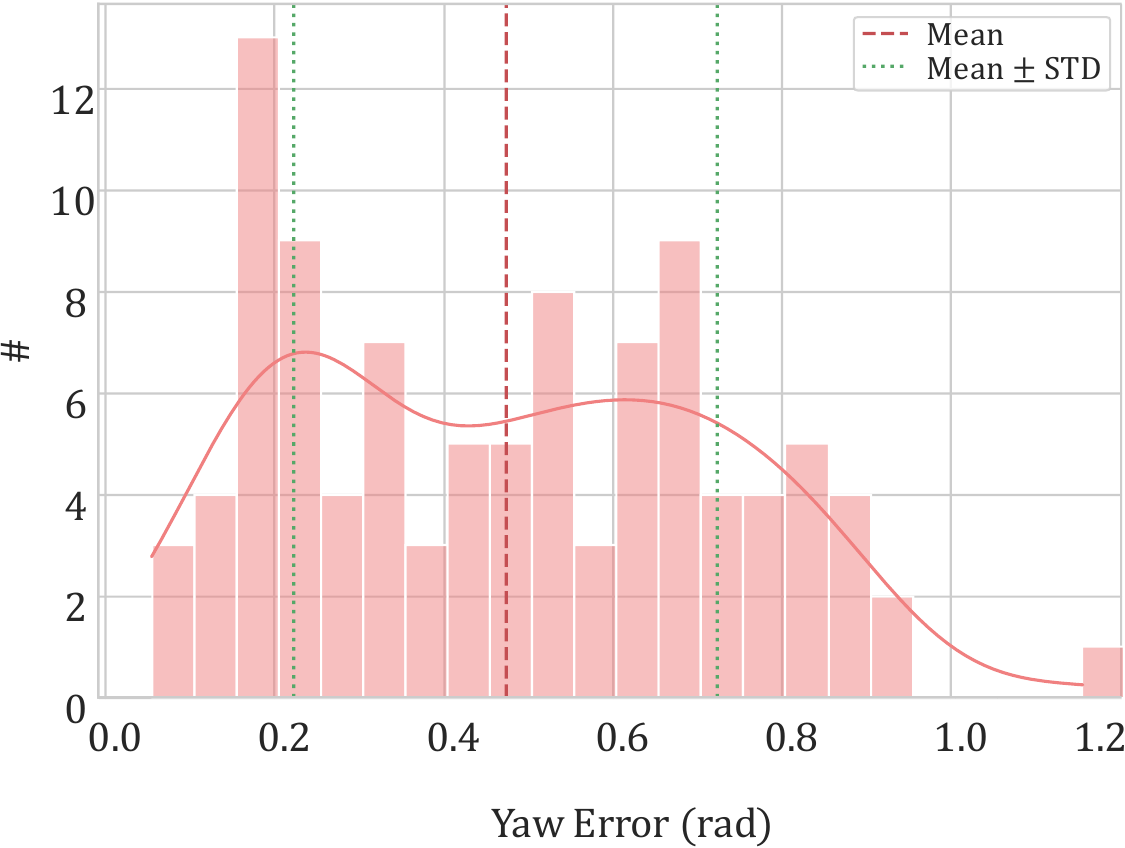}
        \\ (B)
    \end{minipage}

    \caption{Error between the final pose and the pose where the desired target image was captured in open-loop simulations: (A) position; (B) yaw.}
    \label{fig:final_error}
\end{figure*}

To give an intuition of the probabilistic nature of the proposed approach, a batch of 100 open-loop trajectories generated by the DDPM is shown in~\figref{fig:open_loop}, along with the highlight of the initial target-blind view and of a sample successful final view. We analyzed the residual error between the final pose and the pose where the desired target view was captured.~\figref{fig:final_error} shows the distribution of the residual error in the form of histograms.

\subsection{DDPM-based controller}

\begin{table}[t]
    \caption{Estimated return values at different consecutive samplings of the receding horizon control in the two simulated scenarios.}
    \label{tab:return}
    \centering
    \begin{tabular}{lcccccc}
        \toprule
        Scenario & \multicolumn{6}{c}{Return} \\
        \midrule
                 & 1    & 2    & 3    & 4    & 5    & 6    \\
        Box      & 0.62 & 0.69 & 0.71 & 0.76 & 0.87 & 0.88 \\
        Lab      & 0.46 & 0.52 & 0.60 & 0.64 & 0.68 & 0.73 \\
        \bottomrule
    \end{tabular}
\end{table}

In the following, we analyze the performance of the proposed approach, showing that the receding horizon paradigm is the key to reaching the visual target and that the return estimation algorithm consistently estimates lower returns the more the robot approaches the goal.

In~\figref{fig:motivation}, we have shown the closed-loop experiment for the hexarotor in a simulated lab environment. The DDPM samples a trajectory, then the first five control actions are executed before sampling a new trajectory again. As shown in~\tabref{tab:return}, the estimated return is consistently growing as the UAV approaches the target, both in the lab and in the box simulation environments. As it is well known, the closed-loop experiment allows the correction of the uncertainties in the model. As a consequence, the average residual error can be drastically reduced: on one of the initial positions, it improved from 1.22 m to 0.10 m in position, and from 0.43 rad to 0.40 rad in yaw.

Further, applying the latent DDPM in a receding horizon fashion allows (i) sampling shorter trajectories, thus reducing the sampling time, and (ii) successfully using low-quality datasets where, e.g., the trajectories are short, sub-optimal, or do not reach the target at all.
%%%%%%%%%%%%%%%%%%%%%%%%%%%%%%%%%%%%%%%%%%%%%%%%%%%%%%%%%%%%%%%%%%%%%%%%%%%%%%%%
\section{Conclusion}
\label{sec:conclusion}

In this paper, we presented a novel VS approach based on latent DDPM, capable of handling target-invisible views. It relies on coupling a CM-VAE for the latent representation and DDPMs for planning latent space trajectories. Additionally, we introduced a heuristic to estimate the desired return to be conditioned on that drastically improves the generated trajectories by the DDPM. 

In the current implementation, the trajectory generated by the DDPM is followed by relying on a Motion Capture system for localization. As future work, we aim to remove the dependency on a localization pipeline by developing a full end-to-end approach. This way, the invisible servoing algorithm will overcome the limitations of traditional localization pipelines, which struggle in case of lack of features or bad lighting. Indeed, such challenging conditions can be reproduced in the dataset and the models trained to guarantee successful servoing. To achieve the objective of an end-to-end pipeline, the sampling process needs to be sped up. The following strategies will be considered: (i) using more powerful hardware, e.g., Nvidia Jetson, and reducing the diffusion steps using (ii) warm-start strategies, i.e., exploiting previously-generated trajectories as initial guesses for the DDPM to denoise the next trajectory, and (iii)  leveraging the research on novel and computationally efficient DDPMs, such as the Denoising Diffusion Implicit Models \cite{song2020denoising}. Future work also includes combining the DDMP approach with traditional MPC. This would allow including additional objectives and dynamic feasibility of the solution.

In addition, the approach will be experimentally validated in real-world and more complex demonstration scenarios.

%%%%%%%%%%%%%%%%%%%%%%%%%%%%%%%%%%%%%%%%%%%%%%%%%%%%%%%%%%%%%%%%%%%%%%%%%%%%%%%%
\section*{Acknowledgments}
The authors thank Anthony Mallet (LAAS-CNRS) for the camera driver software support, along with all the contributors to the Aerial Robotics Testbed project\footnote{\url{https://git.openrobots.org/projects/art}}.
\balance

%Bibliography
\bibliographystyle{unsrt}  
\bibliography{references}

\end{document}